\RequirePackage{amsmath}
\documentclass[runningheads]{llncs}
\usepackage{graphicx}
\usepackage{mathtools}
\usepackage{subfigure}
\usepackage{epstopdf}
\AppendGraphicsExtensions{.tif}

\usepackage{caption}
\usepackage{float} 
\usepackage{hyperref}
\hypersetup{hidelinks,hypertexnames=false}
\usepackage{cleveref}

\usepackage[ruled,vlined]{algorithm2e}
\usepackage{todonotes}

\usepackage{bookmark}

\usepackage{tabularx}
\usepackage{array}
\usepackage{hhline}
\newcolumntype{M}[1]{>{\centering\arraybackslash}m{#1}}
\usepackage{longtable}
\usepackage{multirow}

\usepackage{booktabs}

\usepackage{amsmath}
\usepackage{amsfonts}

\usepackage[misc,geometry]{ifsym}

\usepackage{xcolor}

\usepackage{lscape} 

\usepackage{enumitem} 

\begin{document}
\title{StaDRe and StaDRo: Reliability and Robustness Estimation of ML-based Forecasting using Statistical Distance Measures}
%
\titlerunning{SafeML - StaDRe and StaDRo}
\author{Mohammed Naveed Akram\inst{2}\thanks{Corresponding author: email: naveed.akram@iese.fraunhofer.de} \and Akshatha Ambekar\inst{3} \and Ioannis Sorokos\inst{2}\thanks{Corresponding author: email: ioannis.sorokos@iese.fraunhofer.de} \and Koorosh Aslansefat\inst{1} \and Daniel Schneider\inst{2}}
\authorrunning{MN. Akram et al.}
%
\institute{University of Hull, United Kingdom \and Fraunhofer IESE, Germany \and Technical University of Kaiserslautern, Germany}
\maketitle              
\begin{abstract}

Reliability estimation of Machine Learning (ML) models is becoming a crucial subject. This is particularly the case when such \mbox{models} are deployed in safety-critical applications, as the decisions based on model predictions can result in hazardous situations. As such, recent research has proposed methods to achieve safe, \mbox{dependable} and reliable ML systems. One such method is to detect and analyze distributional shift, and then measuring how such systems respond to these shifts. This was proposed in earlier work in SafeML. This work focuses on the use of SafeML for time series data, and on reliability and robustness estimation of ML-forecasting methods using statistical distance measures. To this end, distance measures based on the Empirical Cumulative Distribution Function (ECDF), proposed in SafeML, are explored to measure Statistical-Distance Dissimilarity (SDD) across time series. We then propose SDD-based Reliability Estimate (StaDRe) and SDD-based Robustness (StaDRo) measures. With the help of clustering technique, identification of similarity between statistical properties of data seen during training, and the forecasts is done. The proposed method is capable of providing a link between dataset SDD and Key Performance Indices (KPIs) of the ML models. 

\end{abstract}

\section{Introduction}
\pdfbookmark[section]{Introduction}{sec_intro}
\label{sec_intro}
The improved performance of hardware has triggered the increased use of ML-based components in many applications. This also includes many safety-critical application domains such as medical, industrial, transportation, aviation etc. This in turn requires new methods for increasing and assuring the reliability of ML-based components in safety-critical applications. One way to achieve this increase is by detecting dataset shift \cite {storkey2009training}. In previous work, SafeML has been introduced by the authors in \cite{aslansefat2020safeml}, which uses statistical distance measures based on the Empirical Cumulative Distribution Function (ECDF) to detect distributional shift at runtime. Further, the approach estimates the performance degradation of a classifier based on those measures. The results showed that there is a correlation between performance and Statistical Distance Dissimilarity (SDD). The proposed method, however, was limited to only tabular data and classification tasks. In \cite{aslansefat2021toward}, SafeML has been improved to handle image classification tasks. Also, the authors have proposed to use a bootstrap-based p-value calculation to improve the reliability of statistical distance measures. In this work, we extend these measures for a time series data and use SDD to estimate reliability (StaDRe), and robustness (StaDRo) of a ML component.

The rest of the paper is divided as follows. Section \ref{sec_lit} describes related literature, whereas in section \ref{sec_problem} the problem statement is discussed. Section \ref{sec_methodology} details the proposed methodology, and section \ref{sec_experiments} describes the experiments conducted. Section \ref{sec_result} discusses the results, concluding the paper in \ref{sec_conclusion}.


\section{Related Works}
\label{sec_lit}
\pdfbookmark[section]{Related Works}{sec_lit}

For an ML-based system to be used in safety-critical systems, certain qualities of ML solutions such as reliability, fairness, robustness, transparency and security are important. Several methods have been proposed to quantify and qualify these properties. For this work, the properties of ML-based system resulting from the dataset distribution and shifts are considered.

This section is divided into several subsections. Section \ref{lit_data_shift} discusses the available methods for detecting dataset shift and distributional shift. Section \ref{lit_reliability} discusses the existing reliability metrics available in the literature. And section \ref{lit_robustness} discusses the existing robustness evaluation methods. 

\subsection{Dataset shift}
\label{lit_data_shift}
\pdfbookmark[subsection]{Dataset shift}{lit_data_shift}
There exist several names for the concept of dataset shift in the literature. Some of these are concept  shift, contrast mining (in classification  learning, fracture between data and fracture points and changing environments \cite{moreno2012unifying}). To overcome the inconsistencies, authors in \cite{moreno2012unifying} have defined dataset shift as the difference in the joint distribution in training and runtime datasets. Additionally, they identify three types of dataset shift: A covariate shift, prior probability shift, and a concept shift. 

Several techniques for dataset shift detection are proposed in the literature for forecasting and classification problems. The work in \cite{BECKER20213391} proposes measures that are applicable only to linear and multiple linear regression models, work on exponentially-weighted moving average (EWMA) charts \cite{raza2015ewma} is limited by analysis power in presence of larger shifts, the proposed technique in \cite{rabanser2019failing} for image classification problem is restricted by lack of support for online dataset shift detection and the technique based on Particle Swarm Optimization (PSO) \cite{oliveira2017time} overshadows its usefulness due to an overhead of retraining every time a concept drift occurs. In addition to these limitations, the existing methods do not consider statistical distance measures for computing distributional shift in time series.

\subsection{Reliability}
\label{lit_reliability}
\pdfbookmark[subsection]{Reliability}{lit_reliability}
A widely regarded work in dependable systems, the authors in \cite{avizienis2004basic}, define reliability as continuity of correct service. In classical systems, quality measures such as Mean Time Between Failures (MTBF) is usually used for reliability measurement. For an ML system, authors in \cite{bosnic2009overview} define reliability by any qualitative attribute related to a ``vital performance indicator" of that system. They included measures such as accuracy or inaccuracy, availability, downtime rate, responsiveness etc. In the context of ML-model-driven systems, we consider continuity of correct service, or ML performance evaluation, to not only encompass the service provision, but also correctness of the ML model response itself.

The existing approaches proposed in literature for reliability estimation can be classified into two classes: model-agnostic or model-specific approaches. While model-specific techniques \cite{Gammerman,Nouretdinov} are based on specific model designs and learning algorithms, they utilize probabilistic interpretations for reliability, in terms of confidence measurement. In our work, we focus on model-agnostic techniques.

The model-agnostic approaches are generic and provide distributional or indicator-based reliability estimates that are less probabilistically interpretable. Some examples of model-agnostic reliability estimation techniques are sensitivity analysis \cite{bosnic2008estimation}, local cross-validation \cite{demut2010reliability}, confidence estimation based on neighbors’ errors and
variance in the environment \cite{briesemeister2012no}, bootstrapping \cite{efron1992bootstrap} and an ML-based approach \cite{adomavicius2021improving}. Authors in \cite{briesemeister2012no} propose the  \textit{``CONFINE"} metric to measure  reliability. In contrast to previous heuristic-based techniques, the CONFINE measure incorporates actual model prediction errors for reliability estimation, and estimations are based on the error of the test instance's nearest neighbors. However, this metric does not cover the effect of distributional shift on reliability. Hence, we propose extending the CONFINE metric with the rate of change of dissimilarity.

\subsection{Robustness}
\label{lit_robustness}
\pdfbookmark[subsection]{Robustness}{lit_robustness}

In machine learning literature, different aspects of model's robustness such as robustness against adversarial attacks and robustness against dataset shift are evaluated. To begin, adversarial robustness is used to measure the robustness of an ML model against adversarial attacks. The authors of \cite{rauber2017foolbox} describe model robustness as the average size of the minimum adversarial perturbation over many samples. The majority of work in this field focuses on various kinds of adversarial attacks and building ML models that are robust against these.

However, in this work, we focus on robustness against dataset shift. A well-generalized network should be robust to data subpopulations it has not yet encountered \cite{santurkar2020breeds}. A measure termed \textit{relative accuracy} to compute model’s performance is presented in work \cite{santurkar2020breeds} with a focus on image classification problems. Furthermore, the work in \cite{hendrycks2019augmix} argues that accuracy can suffer when the train and test distributions are dissimilar, therefore implement a data augmentation technique at training to improve robustness of image classifiers, as well as use \mbox{\textit{flip probability}} as a measure of perturbation robustness. The authors of~\cite{taori2020measuring} \mbox{introduced measure} called \textit{effective robustness} which is the model accuracy difference in the presence of distributional shift, taking into consideration an accuracy baseline determined using available standard models (without robustness interventions). The authors of \cite{aslansefat2020safeml} examined the use of statistical distance measures to evaluate the robustness of ML models. The work in \cite{hond2021integrated} propose verification of ML models in safety-critical applications, by verifying generalized model performance using Neuron Region Distance (NRD) as an aid for dissimilarity measure.

A ML model that is robust should be able to deliver acceptable performance under vulnerabilities like dataset shift. This notion of robustness is also closely related to definition of robustness presented in \cite{IEEEGlossary} and \cite{avizienis2004basic} for classical systems. Though the existing methods cover a plethora of robustness measures, currently there are not many that use statistical distance measures to account for robustness against dataset shift. In our work, we propose using a performance vs dissimilarity curve for robustness evaluation.

\section{Problem Definition}
\label{sec_problem}
\pdfbookmark[section]{Problem Definition}{sec_problem}
%

When an ML model encounters an input differing from its training data, out-of-distribution (OOD) uncertainty emerges. This can be taken into account by estimating the SDD, with which reliability and robustness can be estimated. Hence, in this work, the following research questions are established:

\begin{enumerate}[leftmargin=30pt]
    \item[\textbf{RQ1:}] \textbf{SDD-Accuracy Correlation}. How does SDD relate to the performance of a model specifically for time series application?
    \item[\textbf{RQ2:}] \textbf{SDD based Reliability and Robustness}. How can SDD be incorporated in reliability and robustness measures of a model?
\end{enumerate}

\section{Proposed Method}
\label{sec_methodology}
\pdfbookmark[section]{Proposed Method}{sec_methodology}


In this section,
we use the SDD measure to obtain the StaDRe and StaDRo. Finally, we describe the experimental design for our evaluation of the proposed approaches on a univariate time series dataset.

For the rest of the paper, we will use the following terms. Let us assume $X_{all} \in \mathbb{R}$ be the overall univariate time series dataset, $X \in \mathbb{R}$ be the training portion of it, and $Y \in \mathbb{R}$ be the validation part. At runtime, we use $X^*$ for then available data. Let $f$ be any ECDF-based distance measure such as Wasserstein, Kolmogorov-Smirnov etc. Let $M$ be the ML model function, and $Y' \in \mathbb{R}$ be the prediction for a given input $x$. Due to its unbounded nature, geometry considerations, and method which is essentially the area measured between two ECDFs, this work primarily uses Wasserstein distance as ECDF function for estimating reliability and robustness \cite{bellemare2017cramer}. 


\subsection{SDD vs Performance}
\label{sec_dissimilarity&Performance}
\pdfbookmark[subsection]{SDD vs Performance}{sec_dissimilarity&Performance}
First, let us consider the behavior of the ML model performance with its SDD against the data observed during training. To observe this behavior, the performance of the validation data $Y$ is compared against its SDD from the training set ($f(X, Y)$). The performance can be measured either in root mean square error (RMSE), or mean absolute percentage error (MAPE). To obtain a sequence of observations, the validation set $Y$ is divided into several subsets $Y^* \in Y$ of length $l$. The performance of the model $M$ can be obtained, by obtaining the prediction over each subset $M(Y^*)$ $\forall{Y^*}$ $\in{Y}$. The corresponding SDD of subset $Y^*$ is obtained using $f(X, Y^*)$. From these, a curve of performance vs SDD can be plotted. Section \ref{sec_experiments} discusses the detailed experiments.

\subsection{Statistical Distance based Reliability estimate - StaDRe}
\pdfbookmark[subsection]{Statistical Distance based Reliability Estimate - StaDRe}{sec_stadre}
In \cite{briesemeister2012no}, reliability of ML model is given in terms of confidence, by ``CONFidence estimation based on the Neighbors’ Errors'' (CONFINE) as in equation \ref{eq_confine}.

\begin{equation}
    \label{eq_confine}
    cs_{CONFINE}(X^*) = 1 - \dfrac{1}{m} \displaystyle \sum_{i=1}^{m} \epsilon_{i}^{2} 
\end{equation}

Where, $X^*$ is a data instance for which reliability is needed, $\epsilon_i$ is the error of a nearest neighbor $i$, $m$ are the total nearest neighbors. However, this does not take into account the SDD of the instance. Moreover, CONFINE uses Euclidean distance to obtain the nearest neighbors.

We extend this approach by taking into account the SDD, and using the DTW-based clustering for computing nearest neighbors of $X^*$. Clustering is a technique commonly used to group together similar patterns in time series using distance measures like DTW, Fr\'echet, Kullback-Leibler divergence \cite{DTWClustering,FrechetClustering,KLclustering}. Among these, K-Means clustering \cite{bradley98} with a DTW distance metric \cite{dtw2004,dtw} was found to be adequate for univariate time series. Due to space limitations, the explanation of the DTW-based clustering technique used is omitted, but is available on GitHub (see \cref{sec_conclusion}).  Instead of using Euclidean distance, we use the DTW-based clustering technique to identify the appropriate cluster $c$ for the input instance $X^*$ obtained from $Y'$. Then, the mean squared error (MSE) is obtained for the $m$ members of the identified cluster $c$. Finally, the ECDF-based distance is computed between the cluster $c$ and  $X^*$, and the cluster $c$ and the reference $O$. The StaDRe is then given by equation \ref{eq_stadre}. The algorithm \ref{alg_stadre} describes the general procedure for StaDRe computation using statistical distance measures. For this work, the following assumptions are made.

\begin{enumerate}
     \item The members of the cluster $c$ to which $X^*$ gets assigned to, are the nearest neighbors of $X^*$, where $X^*$ is a data instance. 
    
    \item To compute $d_{origin}$ in algorithm \ref{alg_stadre} initialised for Wasserstein distance, the origin is a time series containing $0$ at every time step and whose shape is the same as the cluster's center. 
\end{enumerate}
\begin{equation}
   StaDRe(X^*) = \dfrac{2 - \dfrac{1}{m} \displaystyle \sum_{i=1}^{m} \epsilon_{i}^{2} - \dfrac{f(c, X^*)}{f(c, O)}}{2}
   \label{eq_stadre}
\end{equation}

Where, $\epsilon_i$ is the error of a given neighbor $i$.

\begin{algorithm}[!ht]
\caption{Algorithm for StaDRe} 
 \label{alg_stadre}
\SetAlgoLined
\KwResult{Reliability}
 X = GetTrainingSet()\;
 $X^*$ = GetDataInstance()\;
 clusters = GetClusters(X)\;
 \For{c in clusters}{
 comparison = CompareClusters(c, $X^*$)\;
 \If{comparison==True}{
  d = getStatisticalDistance(c, $X^*$)\;
  d\_origin = getStatisticalDistance(c,O)\;
  break\;
 }
 }
 MeanSquaredError = GetClusterMeanSquareError(c)\;
 Reliability = $\dfrac{(2 - MeanSquaredError - \dfrac{d}{d\_origin})}{2}$\;
\Return Reliability\;
 
\end{algorithm}

\vspace*{-6.5mm}
\subsection{Statistical Distance based Robustness - StaDRo}
\pdfbookmark[subsection]{Statistical Distance based Robustness - StaDRo}{sec_stadro}
%

In this section, we introduce a method to obtain an evaluation of robustness against distributional shift using statistical distance measures. For this, the performance vs SDD curve obtained in \ref{sec_dissimilarity&Performance} is used. For a required minimum performance $P_{min}$, using the performance vs SDD curve, the corresponding SDD distance for the $P_{min}$ is obtained. For a given data instance $X^*$ for which robustness is required, the ECDF distance measure is obtained against the  instance $X^*$ and the training set $X$. StaDRo for an instance $X^*$ at a given $P_{min}$ is given by the equation \ref{eq_stadro}.


\begin{equation}
    StaDRo(X^*, P_{min}) = \begin{cases}
    True &\text{$\dfrac{f(X, X^*)}{d_{P{min}}} \leq 1$}\\\\
    False &\text{$\dfrac{f(X, X^*)}{d_{P_{min}}} > 1$}
    \end{cases}
    \label{eq_stadro}
\end{equation}

Where, $StaDRo(X^*, P_{min})$ gives the robustness of the model for the instance $X^*$ at a given minimum required performance $P_{min}$. The algorithm \ref{alg_stadro} shows a general procedure for StadRo estimation using statistical distance measures.

\begin{algorithm}[!ht]
 \caption{Algorithm for StaDRo} 
 \label{alg_stadro}
\SetAlgoLined
\KwResult{Robust}
 X = GetTrainingSubSet()\;
 Y = GetValidationSet()\;
 \For{Sequence in Y}{
  P = GetPerformance(Sequence)\;
  d = getStatisticalDistance(X, Sequence)\;
  Performance.append(P)\;
  SDD.append(d)\;
 }
 Curved = FitCurve(plot(SDD, Performance))\;
 
 $X^*$ = GetDataInstance()\;
 d\_instance = getStatisticalDistance(X, $X^*$)\;
 
 MINIMUM\_REQUIRED\_PERFORMANCE\_LIMIT = CONSTANT\;
 
 d\_from\_curve = getDistanceForPerformance(Curved, MINIMUM\_REQUIRED\_PERFORMANCE\_LIMIT)\;
 
 Ratio = d\_instance /  d\_from\_curve\;
 \uIf{$Ratio <=  1$}{Robust = True\;}\Else{Robust =False\;}
\Return Robust\;
\end{algorithm}


\vspace*{-6.5mm}
\section{Experiments}
\label{sec_experiments}
\pdfbookmark[section]{Experiments}{sec_experiments}
We performed experiments on stock price prediction application (an example of regression applications) and used stock price data of several companies. The closing price of these stocks is considered. We extracted this using the National Stock Exchange (NSE) library and Yahoo Finance. We excluded the dataset that did not exhibit any kind of shift between data at training and validation. The data was divided into 80\% training data and 20\% was used for the validation. We predicted the stock closing price for the next day, at a given point the model is used. All the data was normalized between -1 and 1. The two forecasting models we implemented are Long Short-Term Memory (LSTM) \cite{lstm} and Gated Recurrent Unit (GRU) \cite{gru}. The choice of hyperparameters and other information is given in Table \ref{tab_dataset_details}. All the models are trained using the Adam optimizer \cite{kingma2014adam} with a learning rate of 0.01 over 100 epochs for LSTM and 50 epochs for GRU.

To obtain StaDRo as described in Algorithm \ref{alg_stadro}, the experiments are divided into three phases. In the first phase, we measure how accuracy $acc$ of the model $M$ changes with shift in the subset of the validation set $Y$. In the second phase, $f(X, Y)$ is computed for the same subsets of $Y$ as in the first phase. For this work, we instantiated Wasserstein distance as an ECDF measure $f$. In the third phase, we fit a polynomial curve of degree 2 to derive the relationship between dataset SDD and the forecasting model’s performance. We then obtain a statistical distance for a desired minimum performance from the fitted curve. For a desired sequence $X^*$, we compute the SDD between $X^*$ and $X$, then calculate the ratio and estimate robustness.

For reliability estimation using the general procedure described in Algorithm \ref{alg_stadre}, we applied clustering on training data of several companies and then selected some validation points randomly at different time steps to compute StadRe for these points. When computing MSE, we chose the number of nearest neighbors ($m$) as equal to the number of samples of $X$ that get assigned to the smallest cluster $c'$. We applied random sampling for selecting $m$ neighbors when the assigned cluster $c$ is not $c'$.


\vspace*{-5mm}
\begin{table} 
\centering
\caption{Details of several stocks and hyperparameters of LSTM and GRU model.}
\vspace*{2mm}
\label{tab_dataset_details}
\resizebox{\textwidth}{!}{%
\begin{tabular}{|c|ccccc|}
\hline
\multirow{2}{*}{\textbf{Characteristic}} &
  \multicolumn{5}{c|}{\textbf{Dataset}} \\ \cline{2-6} 
 &
  \multicolumn{1}{c|}{\textit{Reliance}} &
  \multicolumn{1}{c|}{\textit{Google}} &
  \multicolumn{1}{c|}{\textit{Airtel}} &
  \multicolumn{1}{c|}{\textit{JP Morgan}} &
  \textit{MRF} \\ \hline
\textit{ Start date (dd.mm.yyyy) } &
  \multicolumn{1}{c|}{04.01.2010} &
  \multicolumn{1}{c|}{03.01.2005} &
  \multicolumn{1}{c|}{04.01.2010} &
  \multicolumn{1}{c|}{02.01.2003} &
  03.01.2005 \\ \hline
\textit{No. of Data points} &
  \multicolumn{1}{c|}{2895} &
  \multicolumn{1}{c|}{4300} &
  \multicolumn{1}{c|}{3008} &
  \multicolumn{1}{c|}{4803} &
  4214 \\ \hline
\textit{Window size} &
  \multicolumn{1}{c|}{20} &
  \multicolumn{1}{c|}{50} &
  \multicolumn{1}{c|}{20} &
  \multicolumn{1}{c|}{20} &
  20 \\ \hline
\textit{No. of clusters} &
  \multicolumn{1}{c|}{7} &
  \multicolumn{1}{c|}{5} &
  \multicolumn{1}{c|}{6} &
  \multicolumn{1}{c|}{5} &
  5 \\ \hline
\textit{\begin{tabular}[c]{@{}c@{}}LSTM\\ (\# layers; hidden size)\end{tabular}} &
  \multicolumn{1}{c|}{2; 32} &
  \multicolumn{1}{c|}{2; 40} &
  \multicolumn{1}{c|}{2; 32} &
  \multicolumn{1}{c|}{2; 64} &
  2; 32 \\ \hline
\textit{\begin{tabular}[c]{@{}c@{}}GRU\\ (\# layers; hidden size)\end{tabular}} &
  \multicolumn{5}{c|}{2; 32} \\ \hline
\end{tabular}%
}
\end{table}
\vspace*{-5mm}

\section{Results and Discussion}
\label{sec_result}
\begin{figure}[!ht]
    \centering
    \includegraphics[width=120mm]{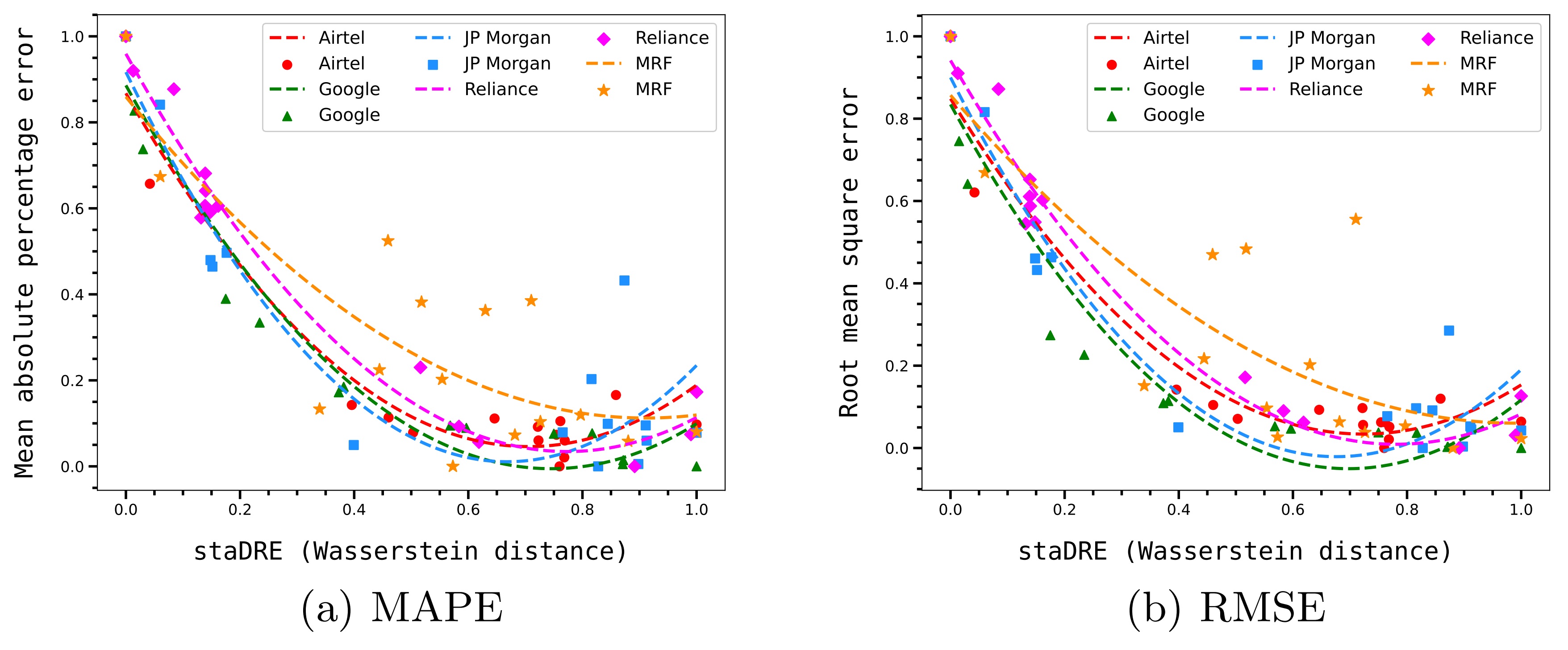}
    \caption{The curve displaying performance vs StaDRe for several stocks normalized to 0-1 range.}
    \label{fig_stadre}
\end{figure}

\pdfbookmark[section]{Results and Discussion}{sec_result}
\begin{figure}[!ht]
    \centering
    \includegraphics[width=120mm]{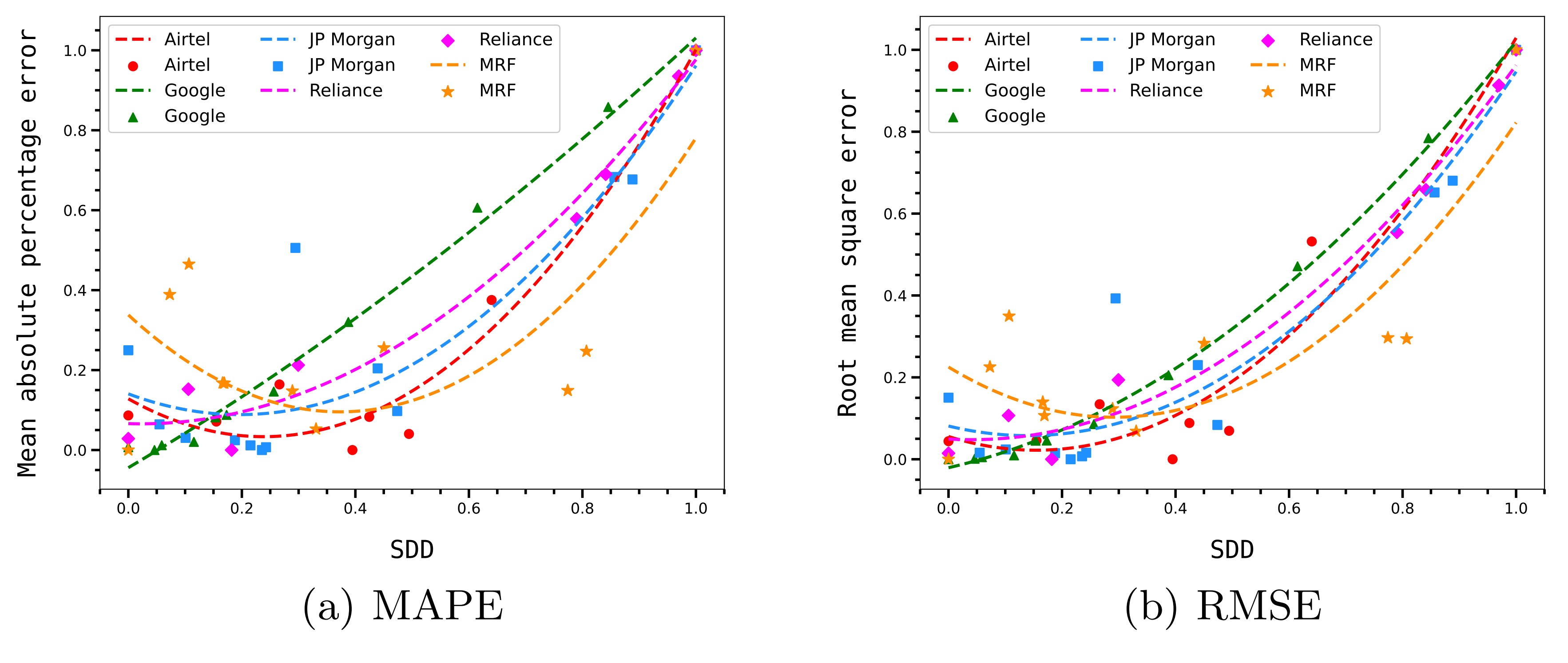}
    \caption{The curve displaying performance vs SDD for several stocks normalized to 0-1 range. }
    \label{fig_wd}
\end{figure}
This section presents the results of the experiments performed. Figure \ref{fig_wd} shows how model performance (as MAPE, RMSE) deteriorates as SDD increases. Each performance measure is fit to a curve using  second-degree curve fitting. As clearly seen in Figures \ref{fig_wd}-a and \ref{fig_wd}-b, there is an increase in performance error with increasing SDD. Figure \ref{fig_stadre} displays the reliability measure against model performance. As with Figure \ref{fig_wd}, the plots are fitted with a second-degree curve. The figures clearly shows increasing performance with increasing reliability, StaDRe, thereby validating the measure.




Table \ref{tab_stadro} shows the results of the StaDRo computation on some example stocks. The table consists of a sequence of validation sets across 2 stocks, corresponding performance metrics (RMSE, MAPE) and StaDRo robustness results for each metric. It can be seen from the table that for a validation subset which StaDRo identifies as not robust against SDD (listed as `FALSE'), a higher error is observed. This indicates that StaDRo is a useful measure for robustness against statistical dissimilarity in datasets.


\vspace*{-5mm}

\begin{table}[ht]
\centering
\caption{Results displaying StaDRo estimates on some stocks}
\vspace*{3mm}
\label{tab_stadro}
\resizebox{\textwidth}{!}{%
\begin{tabular}{|c|cccc|cccc|}
\hline
 &
  \multicolumn{4}{c|}{\textbf{Google ($P_{min}$ (RMSE) = 500.0)}} &
  \multicolumn{4}{c|}{\textbf{JP Morgan ($P_{min}$ (RMSE) = 8.5)}} \\ \cline{2-9} 
\multirow{-2}{*}{\textbf{\begin{tabular}[c]{@{}c@{}}Data \\   Instance\end{tabular}}} &
  \multicolumn{1}{c|}{\textbf{RMSE}} &
  \multicolumn{1}{c|}{\textbf{WD}} &
  \multicolumn{1}{c|}{\textbf{\begin{tabular}[c]{@{}c@{}}Rate of change\\  of SDD\end{tabular}}} &
  \textbf{Robust} &
  \multicolumn{1}{c|}{\textbf{RMSE}} &
  \multicolumn{1}{c|}{\textbf{WD}} &
  \multicolumn{1}{c|}{\textbf{\begin{tabular}[c]{@{}c@{}}Rate of change\\  of SDD\end{tabular}}} &
  \textbf{Robust} \\ \hline
0:70 &
  \multicolumn{1}{c|}{24.85} &
  \multicolumn{1}{c|}{634.73} &
  \multicolumn{1}{c|}{0.39} &
  {\color[HTML]{009901} \textbf{TRUE}} &
  \multicolumn{1}{c|}{2.23} &
  \multicolumn{1}{c|}{61.25} &
  \multicolumn{1}{c|}{0.63} &
  {\color[HTML]{009901} \textbf{TRUE}} \\ \hline
70 : 140 &
  \multicolumn{1}{c|}{31.12} &
  \multicolumn{1}{c|}{740.50} &
  \multicolumn{1}{c|}{0.45} &
  {\color[HTML]{009901} \textbf{TRUE}} &
  \multicolumn{1}{c|}{2.32} &
  \multicolumn{1}{c|}{62.63} &
  \multicolumn{1}{c|}{0.64} &
  {\color[HTML]{009901} \textbf{TRUE}} \\ \hline
140 : 210 &
  \multicolumn{1}{c|}{26.35} &
  \multicolumn{1}{c|}{717.43} &
  \multicolumn{1}{c|}{0.44} &
  {\color[HTML]{009901} \textbf{TRUE}} &
  \multicolumn{1}{c|}{2.53} &
  \multicolumn{1}{c|}{53.44} &
  \multicolumn{1}{c|}{0.55} &
  {\color[HTML]{009901} \textbf{TRUE}} \\ \hline
210 : 280 &
  \multicolumn{1}{c|}{36.73} &
  \multicolumn{1}{c|}{842.07} &
  \multicolumn{1}{c|}{0.52} &
  {\color[HTML]{009901} \textbf{TRUE}} &
  \multicolumn{1}{c|}{2.41} &
  \multicolumn{1}{c|}{59.38} &
  \multicolumn{1}{c|}{0.61} &
  {\color[HTML]{009901} \textbf{TRUE}} \\ \hline
280 : 350 &
  \multicolumn{1}{c|}{82.03} &
  \multicolumn{1}{c|}{910.24} &
  \multicolumn{1}{c|}{0.56} &
  {\color[HTML]{009901} \textbf{TRUE}} &
  \multicolumn{1}{c|}{2.43} &
  \multicolumn{1}{c|}{63.12} &
  \multicolumn{1}{c|}{0.65} &
  {\color[HTML]{009901} \textbf{TRUE}} \\ \hline
350 : 420 &
  \multicolumn{1}{c|}{82.46} &
  \multicolumn{1}{c|}{945.59} &
  \multicolumn{1}{c|}{0.58} &
  {\color[HTML]{009901} \textbf{TRUE}} &
  \multicolumn{1}{c|}{3.28} &
  \multicolumn{1}{c|}{78.90} &
  \multicolumn{1}{c|}{0.81} &
  {\color[HTML]{009901} \textbf{TRUE}} \\ \hline
420 : 490 &
  \multicolumn{1}{c|}{133.10} &
  \multicolumn{1}{c|}{1094.77} &
  \multicolumn{1}{c|}{0.67} &
  {\color[HTML]{009901} \textbf{TRUE}} &
  \multicolumn{1}{c|}{7.16} &
  \multicolumn{1}{c|}{66.65} &
  \multicolumn{1}{c|}{0.68} &
  {\color[HTML]{009901} \textbf{TRUE}} \\ \hline
490 : 560 &
  \multicolumn{1}{c|}{284.06} &
  \multicolumn{1}{c|}{1330.72} &
  \multicolumn{1}{c|}{0.82} &
  {\color[HTML]{009901} \textbf{TRUE}} &
  \multicolumn{1}{c|}{4.12} &
  \multicolumn{1}{c|}{46.56} &
  \multicolumn{1}{c|}{0.48} &
  {\color[HTML]{009901} \textbf{TRUE}} \\ \hline
560 : 630 &
  \multicolumn{1}{c|}{620.01} &
  \multicolumn{1}{c|}{1739.06} &
  \multicolumn{1}{c|}{1.07} &
  {\color[HTML]{FE0000} \textbf{FALSE}} &
  \multicolumn{1}{c|}{2.43} &
  \multicolumn{1}{c|}{50.30} &
  \multicolumn{1}{c|}{0.52} &
  {\color[HTML]{009901} \textbf{TRUE}} \\ \hline
630 : 700 &
  \multicolumn{1}{c|}{1014.97} &
  \multicolumn{1}{c|}{2153.03} &
  \multicolumn{1}{c|}{1.32} &
  {\color[HTML]{FE0000} \textbf{FALSE}} &
  \multicolumn{1}{c|}{5.11} &
  \multicolumn{1}{c|}{76.55} &
  \multicolumn{1}{c|}{0.78} &
  {\color[HTML]{009901} \textbf{TRUE}} \\ \hline
700 : 770 &
  \multicolumn{1}{c|}{1287.46} &
  \multicolumn{1}{c|}{2430.59} &
  \multicolumn{1}{c|}{1.49} &
  {\color[HTML]{FE0000} \textbf{FALSE}} &
  \multicolumn{1}{c|}{10.40} &
  \multicolumn{1}{c|}{105.03} &
  \multicolumn{1}{c|}{1.08} &
  {\color[HTML]{FE0000} \textbf{FALSE}} \\ \hline
770 : 840 &
  \multicolumn{1}{c|}{-} &
  \multicolumn{1}{c|}{-} &
  \multicolumn{1}{c|}{-} &
  - &
  \multicolumn{1}{c|}{10.76} &
  \multicolumn{1}{c|}{107.21} &
  \multicolumn{1}{c|}{1.1} &
  {\color[HTML]{FE0000} \textbf{FALSE}} \\ \hline
840 : 910 &
  \multicolumn{1}{c|}{-} &
  \multicolumn{1}{c|}{-} &
  \multicolumn{1}{c|}{-} &
  - &
  \multicolumn{1}{c|}{14.77} &
  \multicolumn{1}{c|}{114.84} &
  \multicolumn{1}{c|}{1.18} &
  {\color[HTML]{FE0000} \textbf{FALSE}} \\ \hline
\end{tabular}%
}
\end{table}

\vspace*{-2mm}
\section{Conclusion and Future Works}
\label{sec_conclusion}
\pdfbookmark[section]{Conclusion and Future Works}{sec_conclusion}
In this paper, we describe the use of SafeML for SDD-based assessment of ML models for time series applications. The performance behavior against SDD is investigated. It was found that the performance is correlated to the SDD. We propose SDD-based metrics for reliability as StaDRe and robustness as StaDRo. The results on example applications shows the effectiveness of these measures on univariate time series.

For future work, we are planning to experiment with the proposed measures on multivariate time series for both classification and regression tasks. Adversarial attacks can also be used to consider StaDRo for robustness against adversarial examples as well. The idea of this paper can be extended for videos (image-time-series-based inputs) and tested for autonomous vehicles in CARLA\footnote{www.carla.org} simulation. Moreover, the proposed idea can be used as a foundation for time-series safety-related explainability and interpretability.

All results, code, additional results, additional experiments, and exhaustive explanations will be made available in GitHub\footnote{https://github.com/n-akram/TimeSeriesSafeML} repository.

\vspace*{-2mm}
\section*{Acknowledgements}
This work was supported by the Building Trust in Ecosystems and Ecosystem Component (BIECO) Horizon 2020 Project under Grant Agreement 952702.

%
%

%
%
%
%




\bibliographystyle{splncs}

\newpage
\section*{Appendix}
\section{DTW-based K-Means clustering}

Given a univariate time series dataset, X is divided as $D = {X_1, X_2, . . . ., X_n}$ using a
sliding window approach with a window size w. Then, these training samples are
grouped into clusters $c = {c_1, c_2, .., c_k}$ using K-Means clustering with DTW distance for
computing similarity between two time series. Initially, the process begins with 2 clusters and the corresponding Silhouette score \cite{ROUSSEEUW198753} is computed. The clustering process is repeated for a number of clusters starting from a \textit{min} value to some \textit{max} value along with computation of the Silhouette score for every chosen number of clusters. Following the clustering process and score computation, the clusters and their centroids are plotted and carefully evaluated manually for different patterns and different range of mean values present in the existing data. A manual check for patterns reflected in the produced clusters ensures that prominent patterns, such as
a rapid increase or decrease in trend, are adequately captured by the clustering process.
Clustering process is terminated when the results obtained fulfill manually checked
criteria and retrieve the collected clusters. Algorithm \ref{alg_clustering} describes the general procedure for DTW-based K-Means clustering.\\

\begin{algorithm}
\caption{Algorithm for DTW based K-Means clustering} 
\label{alg_clustering}
\SetAlgoLined
\KwResult{Clusters}
X = GetTrainingSet()\;
cluster, centroid, Silhouette\_score = GetDTWBasedClusters(X, 2)\;
Repeat\;
\For{ClusterNumber in range(min, max, step)}{
 cluster, centroid, Score = GetDTWBasedClusters(X, ClusterNumber, Silhouette\_score)\;
 plot(clusters, centroids)\;
  \uIf{ManualCheckOfClusters(clusters, centroids)}{
     break\;
     }
     \uElseIf{Silhouette\_score != Score}{
        Silhouette\_score != Score\;
        ClusterNumber -= step
     }
 }
\Return Clusters\;
\end{algorithm}

\begin{figure}[h!]
    \centering
    \includegraphics[width=\textwidth, height=8cm]{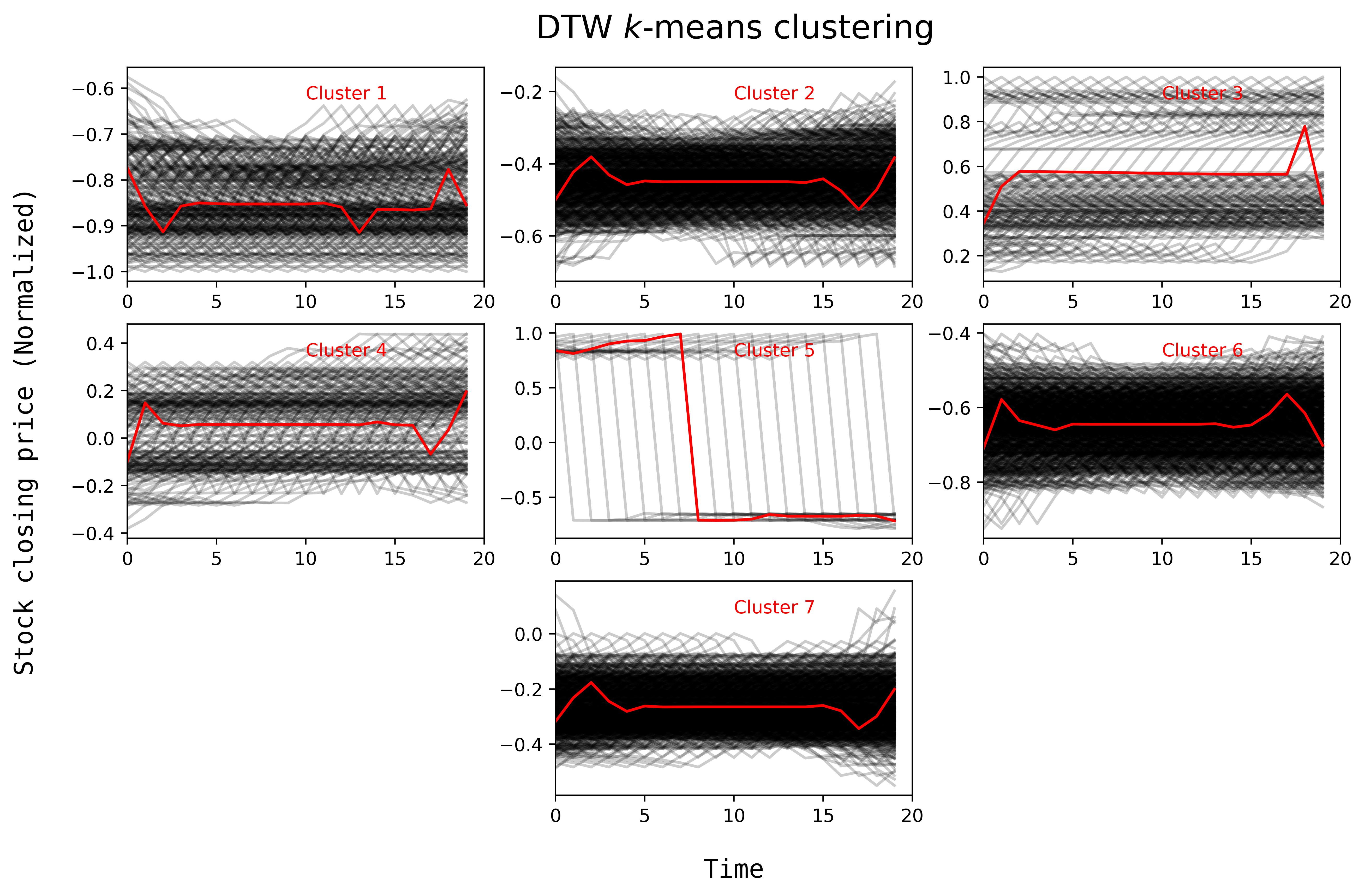}
    \caption{DTW based K-Means clustering for Reliance stock price data.}
    \label{fig_rel_clustering}
\end{figure}

Figure \ref{fig_rel_clustering} visualizes the clusters obtained during the DTW clustering step. In the example shown in Figure \ref{fig_rel_clustering}, there are 7 clusters considered for the Reliance stock price data. We started by selecting 3 clusters that had a better Silhouette score. It was observed that the pattern represented by Cluster 5 in the figure was not captured. Choosing between 5 or 7 clusters did not vary much in terms of Silhouette score, but 7 clusters resulted in a compact range of values (Y-axis) that was desirable when assigning a new test point to a cluster and finding its nearest neighbors.

\end{document}